\documentclass[11pt]{article}

\usepackage[utf8]{inputenc}
\usepackage[T1]{fontenc}
\usepackage{CJKutf8}
\usepackage[a4paper,margin=1in]{geometry}
\usepackage{amsmath}
\usepackage{amssymb}
\usepackage{amsfonts}
\usepackage{graphicx}
\usepackage{xcolor}
\usepackage{booktabs}
\usepackage{multirow}
\usepackage{microtype}
\usepackage{url}
\usepackage[authoryear,round]{natbib}
\usepackage{hyperref}
\hypersetup{colorlinks=true,linkcolor=[rgb]{0.1,0.1,0.5},
            citecolor=[rgb]{0.1,0.1,0.5},urlcolor=[rgb]{0.1,0.1,0.5}}
\microtypesetup{expansion=false}

\graphicspath{{./}}

\title{\textbf{Architecture Shapes Transfer Specificity in Implicit Neural Representations}}
\author{D Yang Eng}

\begin{document}

\maketitle

\begin{abstract}
Transfer in coordinate networks is often measured by warm-start gain, but whether that gain reflects source-specific structure or generic weight reuse is less clear. We study this question across three implicit neural representation (INR) families, SIREN, ReLU MLPs, and Fourier-feature MLPs, using controlled analytic tests, a 2D lid-driven-cavity Navier--Stokes benchmark, and 1D PDE reference-solution suites for heat, viscous Burgers, and focusing cubic NLS. The analytic tests use independent-seed random controls, while the PDE benchmarks use alternate same-family source controls and auxiliary ablations.

Across settings, transfer magnitude and transfer specificity separate clearly. In a 10-seed controlled 1D geometric test, Fourier Features show the largest structured transfer ($33.1\times$), followed by SIREN ($23.0\times$) and ReLU ($10.7\times$), but ReLU is far more selective: random-control transfer is $0.41\times$ for ReLU versus $14.24\times$ for SIREN. On a controlled two-parameter 1D family, the ranking changes: ReLU gives the clearest structured-versus-control separation at default settings, whereas Fourier Features improve only after bandwidth retuning. In Navier--Stokes and the broader 1D PDE suite, no single architecture dominates every equation, yet the same pattern remains: SIREN often reuses weights broadly, whereas ReLU and, in some equations, Fourier Features are more source-selective. Static diagnostics remain weak, and the heuristic scaling law $A_{\text{transfer}} \propto 1/\Delta t^2$ is rejected in the implemented 1D audit.

These results position transfer specificity as a useful diagnostic for coordinate networks and suggest that architecture selection in scientific machine learning should be evaluated under explicit control conditions, not by transfer magnitude alone.
\end{abstract}


\section{Introduction}

Coordinate networks, often called implicit neural representations (INRs), encode continuous signals as neural functions of coordinates. In scientific machine learning, they are attractive as mesh-independent surrogates: once trained, they can be evaluated at arbitrary coordinates, differentiated with respect to their inputs, and reused across steady or time-dependent fields. More broadly, they sit within a scientific-machine-learning push toward reusable surrogate solvers, reduced models, and operator approximations for PDE-governed systems~\citep{BruntonKutz2024}. At the same time, architecture choices induce different spectral biases, frequency preferences, and optimization dynamics, from sinusoidal networks to Fourier-feature MLPs and plain ReLU coordinate networks~\citep{essakine2025,sitzmann2020,Tancik2020,mueller2022}.

Most comparisons between INR architectures focus on reconstruction error or downstream surrogate accuracy. For parametric workflows, however, the more revealing question may be transfer: if a coordinate network has already been trained at one parameter value, when does fine-tuning help at a nearby parameter, and when is that gain specific to the target family rather than generic weight reuse? Framed this way, the problem is also a transfer-learning question about what pretrained neural weights actually carry across related tasks~\citep{PanYang2010}. That question matters in reduced-order and amortized-solver settings, where the point of pretraining is to reuse a learned representation across many related targets rather than fit each instance from scratch.

Here we study transfer behavior itself as the object of interest. We compare SIREN, ReLU MLPs, and Fourier-feature INRs across controlled analytic tests, a 2D Navier--Stokes lid-driven cavity benchmark, and 1D PDE reference-solution suites for heat, Burgers, and focusing cubic NLS. The analytic tests use independent-seed random controls to separate structured transfer from incidental reuse, while the PDE benchmarks use alternate same-family source controls and auxiliary ablations to test whether the same architecture patterns survive in more realistic settings.

Our objective is diagnostic rather than architectural. We do \emph{not} propose a new conditional solver, neural operator, or physics-informed training scheme. Instead, we use transfer magnitude and transfer specificity as complementary probes of representational continuity under a fixed optimization protocol. The central question is: \textbf{how does INR architecture control both the amount of transfer and the source-specificity of that transfer across analytic and PDE families?} The intended output is practical guidance for choosing coordinate-network architectures when transfer, not only single-task accuracy, is the main concern.

\subsection{Key Findings and Contributions}

Our investigation makes four main contributions:

\begin{enumerate}
    \item \textbf{We present a unified transfer study across controlled analytic tests and PDE benchmarks.} The same three INR families are evaluated under explicit control conditions on a 1D geometric family, a controlled two-parameter 1D family, a Navier--Stokes lid-driven cavity benchmark, and 1D PDE reference-solution suites for heat, Burgers, and NLS.

    \item \textbf{Architecture controls transfer specificity as much as transfer magnitude.} In the controlled 1D geometric test, Fourier Features have the largest structured transfer, but ReLU is much more selective under the independent-seed random control, whereas SIREN transfers strongly even on random targets.

    \item \textbf{The architecture ranking is not universal across task families.} ReLU is the clearest discriminator on the controlled two-parameter family and on Navier--Stokes, whereas the heat/Burgers/NLS suite shows that absolute transfer rankings vary by equation even when the distinction between magnitude and specificity persists.

    \item \textbf{Explicit null models matter, whereas simple static diagnostics remain weak.} Independent random controls, alternate-source controls, and shuffled-weight controls materially change the interpretation of transfer gains. By contrast, participation ratio, Hessian sharpness, and independent-seed CKA do not reliably separate structured reuse from weakly specific reuse, and the heuristic scaling law $A \propto 1/\Delta t^2$ is rejected in the implemented 1D audit.
\end{enumerate}

\section{Related Work}

\subsection{Classical and Neural Parametric PDE Surrogates}

\textbf{Reduced-order modeling and classical surrogates.}
Projection-based ROM, POD/Galerkin methods, and reduced-basis approximations are standard tools for repeated parametric PDE solves, especially when many queries are needed after an expensive offline stage~\citep{berkooz1993pod,benner2015survey,quarteroni2015reduced}. Their offline-online structure is conceptually close to the transfer question studied here: one invests in a reusable representation and hopes to reduce the cost of later parameter evaluations. Our INR setting differs in the representation class. Instead of projecting onto a linear reduced basis or a hand-built polynomial basis, we fit coordinate networks to reference fields and ask whether their learned weights provide useful, architecture-dependent warm starts across parameters.

\textbf{Parameter-conditioned PINNs.}
Parameterized and parameter-conditioned PINNs aim to learn continuous PDE solution families by treating physical parameters as network inputs or latent variables. Recent P$^2$INN-style architectures introduce modular or latent-encoded parameter representations that improve accuracy and efficiency on parameterized PDEs~\citep{Cho2024ParameterizedPN,Zhang2025}. Related Navier--Stokes and RANS studies demonstrate both the promise and limitations of parameter-conditioned training for fluid systems~\citep{jangir2026parameterizedphysicsinformedneural,ghosh2023ranspinn}, while GPT-PINN uses a meta-learning strategy that composes pretrained basis PINNs for continuous parameter transfer~\citep{Chen2024}. These methods engineer explicit parameter embeddings to encourage continuity; in contrast, we keep the INR architectures minimal and use transfer behavior across parameters as a probe of how much continuity arises without conditioning.

\textbf{Geometry and domain transfer in PINNs.}
Recent PINN transfer studies include modular fine-tuning schemes with boundary-aware pretraining, lightweight geometry-specific correction layers, and low-rank adaptation across varying boundary conditions, geometries, and material distributions~\citep{li2025modular,Roy_2025,Wang2025tranfer}. Other work encodes irregular geometries into latent variables for physics-informed surrogates or studies transfer through the PINN loss landscape~\citep{Oldenburg2022GAPINN,LIU2023112291}. These studies quantify when transfer is useful across geometries or materials, but they do not compare simple INR coordinate networks across architectures, nor do they use independent-seed random controls to separate genuine target-family continuity from incidental weight reuse.

\textbf{DeepONet and Fourier Neural Operators.}
Operator-learning architectures such as neural operators, DeepONet, and Fourier Neural Operators aim to learn solution operators over whole PDE families rather than fine-tune separate coordinate networks~\citep{Kovachki2023NeuralOperator,lu2021learning,li2021fourier}. Physics-informed DeepONets add PDE residual penalties to learn parametric solution operators without paired input-output data~\citep{Wang2021PID}. FNO variants, including learned-deformation FNOs and factorized FNOs, parameterize solution maps through Fourier-domain kernels and have shown resolution-independent performance on PDE benchmarks~\citep{Li2023,FFNO2021}. More recent operator-learning extensions include physics-informed transformer neural operators for generalized initial/boundary-condition transfer and geometry-aware transformer operators for variable domains~\citep{Boya2024PINTO,Chen2026ArGEnT}. These methods provide powerful alternatives for parametric PDEs; our focus is complementary, asking how much structure is already encoded in simple INR families under naive source-to-target fine-tuning.

\subsection{INR Architectures and Multi-Scale Representation}

\textbf{SIREN, spectral bias, and initialization.}
SIREN introduced periodic activations for INRs and demonstrated that sinusoidal networks are well suited for representing complex signals and their derivatives~\citep{sitzmann2020}. Subsequent work studies how the spectral support and initialization of periodic networks affect optimization. For example, WINNER perturbs uniformly initialized weights using noise scaled by the target signal's spectral centroid, while FINER controls spectral bias through variable-periodic activation functions~\citep{chandravamsi2025,finer}. These results motivate the possibility that SIREN-like networks may reuse weights strongly when their frequency support is aligned with a target family, but may also transfer broadly when that reuse is not specific. The dominance of the architectural prior is not specific to transfer: in a different scientific-machine-learning domain, the choice of geometric prior rather than the coordinate network alone governs accuracy when SIREN-style coordinate networks are used to approximate Calabi--Yau metrics on the quintic~\citep{eng2026cymetrics}, complementing our focus on how architecture shapes transfer rather than single-task accuracy.

\textbf{Fourier features and NTK bandwidth.}
Fourier feature mappings transform the effective neural tangent kernel of an MLP into a stationary kernel with tunable bandwidth, enabling MLPs to fit high-frequency functions that standard ReLU networks learn poorly~\citep{Tancik2020}. This provides a theoretical lens for interpreting architecture-dependent transfer: if reuse depends on spectral alignment, Fourier-feature MLPs may transfer strongly when the feature bandwidth matches the target family, but may lose magnitude or specificity when the bandwidth is mismatched.

\textbf{Multi-resolution and conditional INRs.}
Instant-NGP and related multi-resolution encodings augment coordinate networks with trainable feature grids or hash tables, giving fast optimization and local detail through coarse-to-fine spatial features~\citep{mueller2022,Wang2024,luo2025}. In fluid and PDE settings, Neural Implicit Flow, DINo, conditional neural fields, and related physics-enhanced INRs use hypernetworks, latent dynamics, FiLM-style conditioning, probabilistic residual objectives, or transformer-enhanced coordinate encodings to improve parametric generalization and uncertainty quantification~\citep{NIF,Dino,CNFROM,Ifol,panis,peinr}. Meta-learning has also been used to amortize INR fitting across families of signals, for example through sparse meta-initializations that adapt quickly to unseen targets~\citep{Lee2021MetaSparseINR}. These architectures are explicitly designed to promote cross-instance generalization. Our experiments deliberately use simpler SIREN, ReLU, and Fourier-feature INRs to measure how much transfer magnitude and transfer specificity emerge even without such machinery.

\section{Methods}

\subsection{Controlled Test Families and PDE Benchmarks}

\textbf{Controlled 1D geometric test family:} The family $g_t(x) = \sqrt{x^2 + t^2}$ is an analytic, non-PDE target used to stress-test architecture-dependent transfer near a developing cusp. It is smooth for $t>0$ and approaches $|x|$ as $t \to 0$. This experiment is a controlled diagnostic test, not a physical PDE calculation.

\textbf{Controlled two-parameter 1D test family:} We define a two-parameter \emph{family of 1D target functions} using exponentially damped cosine modes:
\begin{equation}
c_n(x,y) = e^{-\pi n^2 y} \cos(2\pi n x), \quad n = 1, 2, 3
\end{equation}
over the parameter rectangle $(x_{\mathrm{par}},y_{\mathrm{par}}) \in [-0.5, 0.5] \times [0.1, 2.0]$. For each parameter pair we define the 1D signal
\begin{equation}
f_{x_{\mathrm{par}},y_{\mathrm{par}}}(\xi) = c_1(x_{\mathrm{par}},y_{\mathrm{par}})\cos(2\pi \xi)
+ c_2(x_{\mathrm{par}},y_{\mathrm{par}})\cos(4\pi \xi)
+ c_3(x_{\mathrm{par}},y_{\mathrm{par}})\cos(6\pi \xi),
\end{equation}
evaluated on $\xi \in [-1,1]$. This is \emph{not} a modular form or heat kernel on an elliptic curve; it is a smooth analytic family chosen for its multi-scale structure. No modular invariance is assumed or exploited.

To avoid terminological ambiguity, we refer to this experiment as a \emph{two-parameter 1D family} rather than a two-dimensional signal. The parameter space $(x_{\mathrm{par}},y_{\mathrm{par}})$ is two-dimensional, but the represented signal remains a one-dimensional function of $\xi$. Like the 1D geometric family, this is an analytic transfer diagnostic rather than a PDE solve.

\textbf{Random controls for analytic tests:} In both controlled analytic settings, random controls are built by phase-randomizing the Fourier representation of the corresponding structured target while preserving the Fourier magnitudes, then matching the random target to the structured target mean and standard deviation. Source and target random controls use \textbf{independent} seeds, with source seed $s$ and target seed $s+10000$ (plus a direction offset in the two-parameter experiment).

\textbf{PDE benchmarks:} We additionally study three standard one-dimensional PDE families: heat, viscous Burgers, and focusing cubic nonlinear Schrodinger (NLS). These experiments are supervised INR regressions against reference PDE solutions, not physics-informed residual minimization. The heat equation is
\begin{equation}
u_t = \alpha u_{xx}, \qquad x \in [0,1),
\end{equation}
with periodic boundary conditions and initial condition
\begin{equation}
u(x,0)=\sin(2\pi x)+\frac{1}{2}\sin(4\pi x).
\end{equation}
The reference solution is evaluated analytically from the Fourier modes,
\begin{equation}
u(x,t;\alpha)=e^{-\alpha(2\pi)^2t}\sin(2\pi x)
+\frac{1}{2}e^{-\alpha(4\pi)^2t}\sin(4\pi x).
\end{equation}
The source parameter is $\alpha=0.05$, with targets $\alpha \in \{0.02,0.10\}$; the non-designated target parameter is used as the alternate same-family control source.

The Burgers benchmark solves
\begin{equation}
u_t + u u_x = \nu u_{xx}, \qquad x \in [0,1),
\end{equation}
again with periodic boundary conditions and initial condition
\begin{equation}
u(x,0)=\sin(2\pi x)+0.25\sin(4\pi x).
\end{equation}
Reference trajectories are generated with a Fourier pseudo-spectral spatial discretization and classical fourth-order Runge--Kutta time stepping. The source viscosity is $\nu=0.01$, with targets $\nu \in \{0.02,0.005\}$ and the opposite target again serving as the alternate same-family control source.

The NLS benchmark solves the focusing cubic equation
\begin{equation}
i u_t+\frac{1}{2}u_{xx}+\kappa |u|^2 u=0,
\qquad x \in [-10,10),
\end{equation}
with periodic boundary conditions and initial condition $u(x,0)=\mathrm{sech}(x)$. Reference solutions are generated by a Strang split-step Fourier method. The complex-valued solution is represented by two real output channels, $(\mathrm{Re}\,u,\mathrm{Im}\,u)$. The source parameter is $\kappa=1.0$, with targets $\kappa \in \{0.8,1.2\}$ and the opposite target used as the alternate same-family control source. The implementation records the discrete NLS mass drift in the reference metadata as a solver sanity check.

\subsection{Architectures}

We compare three architectures throughout the analytic-test and PDE benchmark studies:
\begin{itemize}
    \item \textbf{SIREN~\citep{sitzmann2020}:} first-layer frequency $\omega_0 = 30$, hidden-layer frequency $\omega_0 = 1$, 3 hidden layers of width 128, with the corresponding SIREN-style initialization used in the implementation
    \item \textbf{ReLU MLP:} 3 hidden layers of width 128 with ReLU activations
    \item \textbf{Fourier Features~\citep{Tancik2020}:} 64 random Fourier frequencies with $\sigma = 10$, followed by the same 3-layer width-128 ReLU MLP
\end{itemize}

These settings define the \emph{default} architecture comparison reported in the main two-parameter 1D family table and figures. In a fairness-oriented re-analysis of the two-parameter experiment, we additionally sweep SIREN first-layer frequency $\omega_0$ and a global SIREN initialization scale, and we sweep ReLU width, depth, and fine-tuning learning rate while keeping the remaining protocol fixed. In the current implementation, both analytic-test and Navier--Stokes Fourier embeddings use the Tancik-style convention $\gamma(v)=[\sin(2\pi Bv),\cos(2\pi Bv)]$; the analytic test studies use 64 sampled frequencies, whereas the Navier--Stokes benchmark uses 128.

For the 1D PDE reference benchmarks, all architectures map normalized space--time coordinates $(\tilde x,\tilde t)\in[-1,1]^2$ to the PDE state. Heat and Burgers use one scalar output channel; NLS uses two real output channels. The reported full-budget PDE runs use width 128 and depth 3 for all architectures, SIREN first-layer frequency $\omega_0=30$, hidden-layer frequency $\omega_0=1$, and Fourier features with $\max(16,\mathrm{width}/2)=64$ sampled frequencies at $\sigma=10$. Reference grids use $N_x=128$ spatial points and $N_t=101$ stored time levels. The quick and smoke profiles use smaller grids and training budgets only for local validation, not for manuscript-scale results.

\subsection{Transfer Learning Protocol}

Transfer advantage is defined as
\begin{equation}
A_{\text{transfer}} = L_{\text{scratch}} / L_{\text{transfer}}
\end{equation}
after $E_{\text{ft}} = 300$ epochs of fine-tuning.

\textbf{Analytic-test training details:} The 1D geometric sweep trains on the full 1000-point grid $x \in [-1,1]$; the two-parameter analytic sweep trains on the full 500-point grid $\xi \in [-1,1]$. All analytic-test runs use Adam with no scheduler, no weight decay, and no early stopping. Pretraining and scratch runs use learning rate $10^{-3}$; fine-tuning uses learning rate $10^{-4}$. Pretraining runs for 1500 epochs and both fine-tuning and scratch training run for 300 epochs. The reported transfer advantage is therefore a fixed-budget final-loss ratio, not a compute-aware metric and not an equal-total-training-cost comparison. Because scratch and fine-tuning do not use learning-rate-matched optimization, these ratios should be interpreted as outcomes of a fixed protocol rather than as optimizer-controlled causal estimates of initialization quality alone. For Fourier features, fine-tuning reuses the pretrained random projection matrix because the whole model is copied, whereas scratch training samples a fresh projection; the reported Fourier transfer therefore combines reuse of the trainable MLP weights with reuse of the fixed random feature map.

\textbf{PDE benchmark training details:} The heat, Burgers, and NLS benchmarks use the same source-transfer versus scratch structure, but with mini-batch training over the fixed space--time reference grid. In the reported full-budget runs, each source or control pretraining run uses 1500 Adam steps with batch size 4096 and learning rate $10^{-3}$. Fine-tuning uses 300 Adam steps with learning rate $10^{-4}$, and scratch training uses the same 300-step budget with learning rate $10^{-3}$. For each PDE, the designated source parameter is fine-tuned to each target parameter. The alternate-source control is not an independent random null; it is the same PDE family pretrained at the opposite target parameter and then fine-tuned to the target of interest. The reported PDE benchmark transfer advantage is computed on the full space--time reference grid as
\begin{equation}
A_{\mathrm{PDE}} =
\frac{\mathrm{MSE}_{\mathrm{scratch}}}
{\mathrm{MSE}_{\mathrm{transfer}}}.
\end{equation}
To isolate optimizer-schedule and Fourier-feature-map effects, we also run two auxiliary PDE controls. First, a matched-optimizer scratch baseline trains scratch models for the same 300-step target budget using the fine-tuning learning rate $10^{-4}$ rather than $10^{-3}$. Second, for Fourier Features, we add two feature-map ablations: one copies the pretrained MLP weights but resamples the fixed random projection matrix $B$ before fine-tuning, and one reuses only the pretrained $B$ while randomly initializing the MLP weights. These controls are reported as diagnostics; the main PDE table keeps the original fixed-budget protocol for comparability with the analytic experiments.
The implementation caches reference solutions by PDE, parameter, grid size, and time-grid size, so rerunning the experiment does not recompute the numerical reference data. The neural training runs themselves are not checkpoint-resumed in the current implementation; the reported full-budget outputs are stored as JSON summaries together with a lightweight SVG diagnostic plot.

The saved revision artifacts also record an auxiliary curve-aware statistic over the full fine-tuning trajectory, $A_{\mathrm{curve}} = \exp\!\big(\mathrm{mean}_t \log((L_{\mathrm{scratch}}(t)+\varepsilon)/(L_{\mathrm{transfer}}(t)+\varepsilon))\big)$, but the headline tables in this manuscript use the terminal ratio $A_{\mathrm{transfer}}$. We keep the terminal ratio as the primary metric because it is directly comparable across all analytic and PDE suites, maps one-to-one to the final target loss after a shared fine-tuning budget, and matches the practical warm-start question of which initialization finishes lower under a fixed target-training budget; $A_{\mathrm{curve}}$ is retained as a robustness check against outlier-sensitive terminal values.

\textbf{Statistical summaries:} The 10-seed 1D geometric sweep and the 10-seed two-parameter analytic sweep report bootstrap percentile 95\% confidence intervals for the mean. In 1D we use one-sided paired Wilcoxon signed-rank tests for within-architecture structured-versus-random comparisons (alternative: structured $>$ random) and two-sided paired Wilcoxon tests for pairwise structured-transfer comparisons across architectures. In the two-parameter family we again use one-sided paired Wilcoxon signed-rank tests for structured-versus-random transfer within each architecture and direction, together with two-sided pairwise architecture comparisons on structured transfer. We report unadjusted $p$-values throughout and therefore use them as localized evidence for pre-specified comparisons rather than as a family-wise error-controlled testing program. As a sensitivity check, a Benjamini--Hochberg correction over the 24 primary Wilcoxon comparisons reported in the main text retains the largest reported passing value at $p=0.020$, which is below the corresponding rank-10 critical value $10 \times 0.05 / 24 \approx 0.0208$; the retained set contains five of the six 1D comparisons (all except the Fourier-versus-SIREN structured comparison), the two-parameter ReLU ``both'' comparison, all three two-parameter Fourier comparisons, and the NS ReLU $100\!\to\!400$ contrast, but not the remaining borderline effects. For the fairness sweeps, because the per-seed ratio $A_{\mathrm{geo}}/(A_{\mathrm{rand}}+10^{-10})$ becomes unstable when $A_{\mathrm{rand}}$ is very small, we interpret those sweeps primarily through macro structured/random means, curve-aware analogues, and the paired Wilcoxon tests rather than through the ratio summary.

\section{Results}

\subsection{Static Diagnostics Remain Weak}

Table~\ref{tab:static} summarizes a cross-architecture diagnostic rerun on the controlled 1D geometric family. The activation-cloud participation ratio, computed on 20 parameter values times 500 spatial samples for a total of 10,000 activation vectors, shows essentially no structured-versus-random separation within any architecture: SIREN gives $94.68$ versus $94.71$, ReLU gives $14.74$ versus $14.98$, and Fourier Features give $15.43$ versus $15.42$. The corresponding parameter-manifold participation ratios are similarly close in each case.

Hessian sharpness, measured as the spectral norm $\lambda_{\max}(H)$ following Foret et al.~\citep{foret2021sharpness}, is likewise weakly informative. Across five seeds, SIREN gives $143.3 \pm 16.5$ for structured targets and $151.4 \pm 17.9$ for random targets, with an exploratory one-sided two-sample $t$-test at $p = 0.375$. ReLU gives $18.7 \pm 0.7$ versus $20.3 \pm 0.8$ with $p = 0.093$, and Fourier gives $14.1 \pm 0.9$ versus $14.1 \pm 0.9$ with $p = 0.500$. We therefore do not treat sharpness as discriminative evidence.

Independent-seed CKA is architecture-dependent, but it is not specificity-selective. On a separate 15-point diagnostics grid, the Mantel Spearman correlation between pairwise CKA distances and pairwise log-parameter distances is $\rho = 0.290$ for SIREN structured targets and $\rho = 0.270$ for SIREN random targets; ReLU gives $\rho = 0.474$ and $\rho = 0.449$, and Fourier gives $\rho = -0.112$ and $\rho = 0.268$. In every architecture, the shared-seed structured baseline is much larger, with $\rho = 0.754$ for SIREN, $\rho = 0.749$ for ReLU, and $\rho = 0.757$ for Fourier. The correct conclusion is therefore not that CKA is always uncorrelated with parameter distance, but that under the valid independent-seed null this Mantel statistic does not cleanly separate structured transfer from random-control transfer.

\begin{table}[t]
\centering
\caption{Cross-architecture static diagnostics on the controlled 1D geometric test family. The participation ratio is computed on the full activation cloud rather than on one mean-pooled vector per parameter value. Sharpness entries are mean $\pm$ standard error over five seeds.}
\label{tab:static}
\small
\resizebox{\textwidth}{!}{%
\begin{tabular}{lccccc}
\toprule
\textbf{Architecture} & \textbf{PR Geo} & \textbf{PR Rand} & \textbf{Sharpness Geo} & \textbf{Sharpness Rand} & \textbf{Sharpness $p$} \\
\midrule
SIREN & $94.68$ & $94.71$ & $143.3 \pm 16.5$ & $151.4 \pm 17.9$ & $0.375$ \\
ReLU MLP & $14.74$ & $14.98$ & $18.7 \pm 0.7$ & $20.3 \pm 0.8$ & $0.093$ \\
Fourier Features & $15.43$ & $15.42$ & $14.1 \pm 0.9$ & $14.1 \pm 0.9$ & $0.500$ \\
\bottomrule
\end{tabular}
}
\end{table}

\subsection{Architecture Dependence in 1D}

\begin{figure}[t]
\centering
\includegraphics[width=0.85\textwidth]{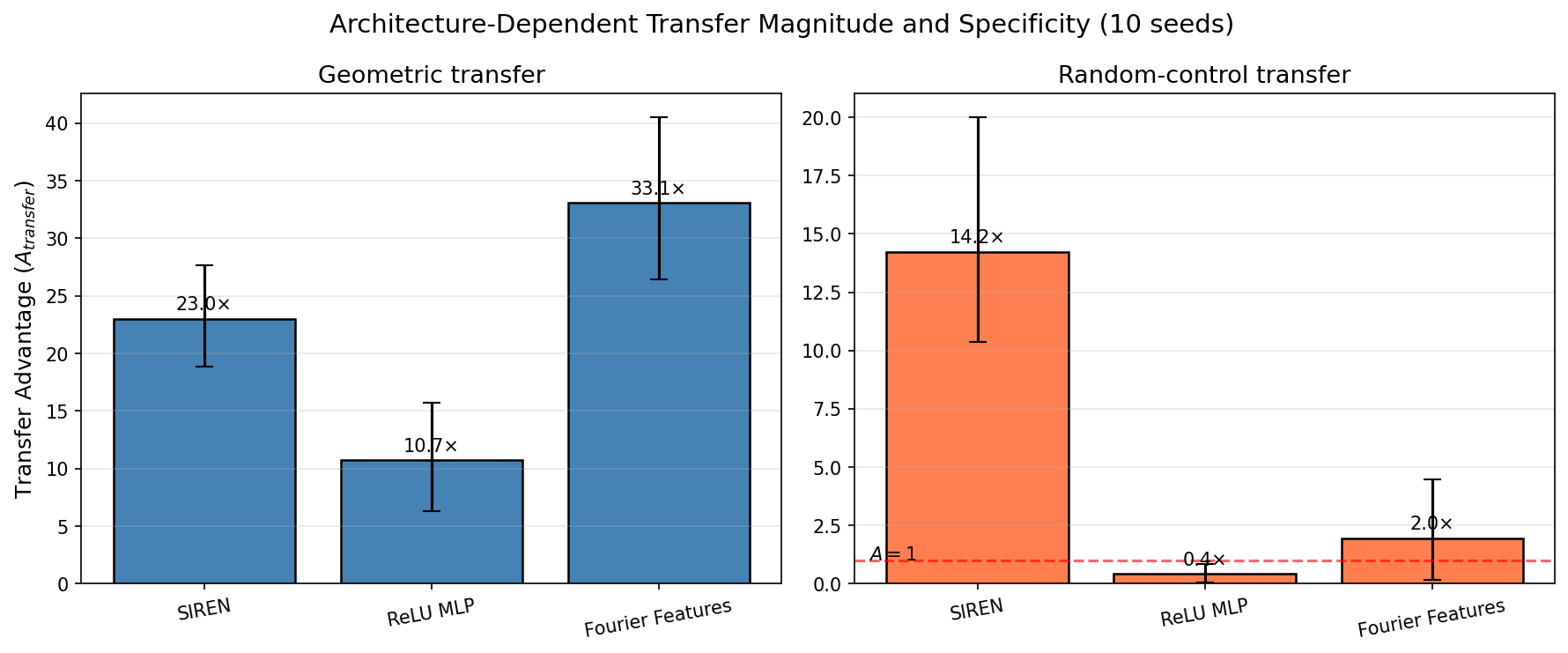}
\caption{Architecture-dependent transfer magnitude and specificity in the controlled 1D geometric test over 10 seeds. Left: structured-target transfer advantage. Right: independent-seed random-control transfer advantage. Error bars denote bootstrap 95\% confidence intervals for the mean. Fourier Features have the largest mean structured transfer, while SIREN shows much less specificity because it also transfers strongly on random targets.}
\label{fig:architecture}
\end{figure}

Table~\ref{tab:arch} and Figure~\ref{fig:architecture} summarize the 10-seed controlled 1D geometric test. Two patterns are clear. First, \emph{absolute structured transfer} differs strongly across architectures: Fourier Features have the largest mean advantage at $33.1\times$, SIREN is intermediate at $23.0\times$, and ReLU is lower at $10.7\times$. Second, \emph{transfer specificity} differs even more sharply: ReLU random-control transfer is close to zero, Fourier random-control transfer is much smaller than its structured transfer but not negligible, and SIREN random-control transfer remains large.

\begin{table}[t]
\centering
\caption{10-seed architecture comparison for the controlled 1D geometric test, $t=0.5 \to 0.55$. Entries are mean transfer advantages with bootstrap 95\% confidence intervals. Wilcoxon $p$-values are from one-sided paired signed-rank tests of whether structured-target transfer exceeds random-control transfer within each architecture.}
\label{tab:arch}
\small
\begin{tabular}{lccc}
\toprule
\textbf{Architecture} & \textbf{Structured} & \textbf{Random} & \textbf{Wilcoxon $p$} \\
\midrule
Fourier Features & $33.1\times\ [26.5, 40.5]$ & $1.96\times\ [0.15, 4.48]$ & $0.0010$ \\
ReLU MLP & $10.7\times\ [6.3, 15.7]$ & $0.41\times\ [0.06, 0.84]$ & $0.0010$ \\
SIREN & $23.0\times\ [18.9, 27.7]$ & $14.24\times\ [10.35, 20.01]$ & $0.0137$ \\
\bottomrule
\end{tabular}
\end{table}

Paired structured-transfer tests still separate both Fourier and SIREN from ReLU, with SIREN exceeding ReLU ($p = 0.0098$) and Fourier exceeding ReLU ($p = 0.0020$). The Fourier-versus-SIREN gap is much weaker after the rerun ($p = 0.0488$). The central issue is not merely that one architecture has a larger transfer number than another. Rather, architecture determines both how large transfer is on the structured target family and how much of that reuse survives a properly independent random null model. ReLU and Fourier show strong structured-vs-random separation; SIREN does not.

\subsection{Two-Parameter 1D Family: Architecture Dependence Changes Character}

Table~\ref{tab:2d_specificity} summarizes the paired 10-seed sweep on the two-parameter 1D family across all three architectures at the default settings. Three different regimes emerge.

First, \textbf{SIREN} again shows substantial structured transfer, but weak specificity: its structured transfer lies between $18.0\times$ and $21.3\times$, while random-control transfer remains between $13.9\times$ and $17.4\times$, and none of the paired Wilcoxon tests reaches the $0.05$ threshold. Second, \textbf{ReLU} shows the clearest specificity in the two-parameter setting. Its structured transfer is large for $x$-only and $y$-only adaptation, and even when both parameters vary it remains well above its random-control baseline; under the pre-specified one-sided within-architecture tests, all three comparisons satisfy $p \leq 0.0322$. Third, \textbf{Fourier Features} are the opposite of their 1D behavior: they still separate structured targets from random controls in a statistical sense, but their absolute structured transfer remains modest, ranging from $0.18\times$ to $1.76\times$ across the three directions.

These results show that the two-parameter story is not a simple extension of the 1D ranking. In this controlled analytic family, explicit Fourier features excel in 1D but remain weak in absolute transfer at the default scale, ReLU becomes the clearest discriminator at the default settings, and default SIREN remains reusable but only weakly specific.

Fairness-oriented within-family sweeps modify the family-level interpretation. For SIREN, the default setting is weakly specific, but reducing the initialization scale from $1$ to $0.5$ changes the macro means from $22.96\times$ structured versus $16.32\times$ random to $20.94\times$ versus $2.91\times$, and all three directional Wilcoxon tests become significant ($p = 0.0010$, $0.0010$, $0.0098$). Increasing $\omega_0$ to $100$ also improves specificity, giving macro means $20.23\times$ versus $10.04\times$ with all three directions significant. The sweep baseline at init-scale $1$ is not numerically identical to the default Table~\ref{tab:2d_specificity} SIREN baseline, because the fairness sweep uses a different architecture-index seed offset; we therefore interpret the sweep as a within-family sensitivity study rather than as a literal re-estimation of the default table. For ReLU, the family is more robust across the tested sweeps: width $128$ remains strong, and depth $4$ is the strongest clean setting among the tested ReLU variants, with macro means $35.87\times$ versus $2.12\times$ and all three directional tests significant ($p \leq 0.0244$). By contrast, some apparently dramatic terminal-ratio wins, such as ReLU depth $5$, are not stable across summaries: the macro terminal advantage rises to $366.94\times$, but the corresponding macro random transfer is still $24.28\times$ and the curve-aware macro advantage is only $19.83\times$, indicating heavy outlier sensitivity rather than a clean family-level improvement. The fairness sweeps therefore support a narrower claim than the default table alone: ReLU remains the most robust two-parameter family across the tested settings, but SIREN is more tunable and more specific under some settings than the default comparison suggests.

\begin{table}[t]
\centering
\caption{Paired 10-seed sweep on the controlled two-parameter 1D family at the default architecture settings. Entries are mean transfer advantages with bootstrap 95\% confidence intervals. Wilcoxon $p$-values are from one-sided paired signed-rank tests of whether structured-target transfer exceeds random-control transfer within each architecture and direction. The ratio column summarizes the per-seed quantity $A_{\mathrm{geo}}/(A_{\mathrm{rand}}+10^{-10})$; it is \emph{not} the ratio of the reported structured and random means. Because $A_{\mathrm{rand}}$ can be very small, this ratio is unstable and is included only as a secondary descriptive statistic.}
\label{tab:2d_specificity}
\small
\resizebox{\textwidth}{!}{%
\begin{tabular}{llcccc}
\toprule
    \textbf{Architecture} & \textbf{Direction} & \textbf{Structured} & \textbf{Random} & \textbf{Per-seed ratio} & \textbf{Wilcoxon $p$} \\
\midrule
\multirow{3}{*}{SIREN} & $x$ only & $18.4\times\ [12.7, 24.0]$ & $13.9\times\ [9.9, 18.0]$ & $1.96\times\ [1.03, 3.00]$ & $0.2158$ \\
 & $y$ only & $21.3\times\ [15.5, 27.7]$ & $15.1\times\ [11.8, 19.9]$ & $1.66\times\ [1.15, 2.17]$ & $0.0654$ \\
 & both & $18.0\times\ [13.7, 23.2]$ & $17.4\times\ [12.5, 23.7]$ & $1.26\times\ [0.85, 1.77]$ & $0.5000$ \\
\midrule
\multirow{3}{*}{ReLU} & $x$ only & $38.6\times\ [14.7, 76.8]$ & $6.9\times\ [0.1, 20.2]$ & $2.59{\times}10^3\ [3.43{\times}10^2, 6.17{\times}10^3]$ & $0.0322$ \\
 & $y$ only & $32.5\times\ [6.2, 79.8]$ & $3.2\times\ [0.0, 9.1]$ & $1.53{\times}10^4\ [6.45{\times}10^2, 4.15{\times}10^4]$ & $0.0322$ \\
 & both & $6.5\times\ [3.6, 10.0]$ & $0.5\times\ [0.1, 1.2]$ & $3.18{\times}10^3\ [6.77{\times}10^1, 8.19{\times}10^3]$ & $0.0010$ \\
\midrule
\multirow{3}{*}{Fourier} & $x$ only & $1.76\times\ [0.74, 2.85]$ & $1.33{\times}10^{-2}\ [1.37{\times}10^{-3}, 3.56{\times}10^{-2}]$ & $1.55{\times}10^3\ [4.94{\times}10^2, 2.84{\times}10^3]$ & $0.0010$ \\
 & $y$ only & $0.18\times\ [0.07, 0.33]$ & $9.90{\times}10^{-4}\ [1.75{\times}10^{-4}, 2.32{\times}10^{-3}]$ & $6.66{\times}10^3\ [3.29{\times}10^2, 1.87{\times}10^4]$ & $0.0010$ \\
 & both & $0.52\times\ [0.02, 1.51]$ & $8.31{\times}10^{-3}\ [8.41{\times}10^{-4}, 2.01{\times}10^{-2}]$ & $2.56{\times}10^4\ [1.65{\times}10^1, 7.66{\times}10^4]$ & $0.0098$ \\
\bottomrule
\end{tabular}
}
\end{table}

\begin{figure}[t]
\centering
\includegraphics[width=\textwidth]{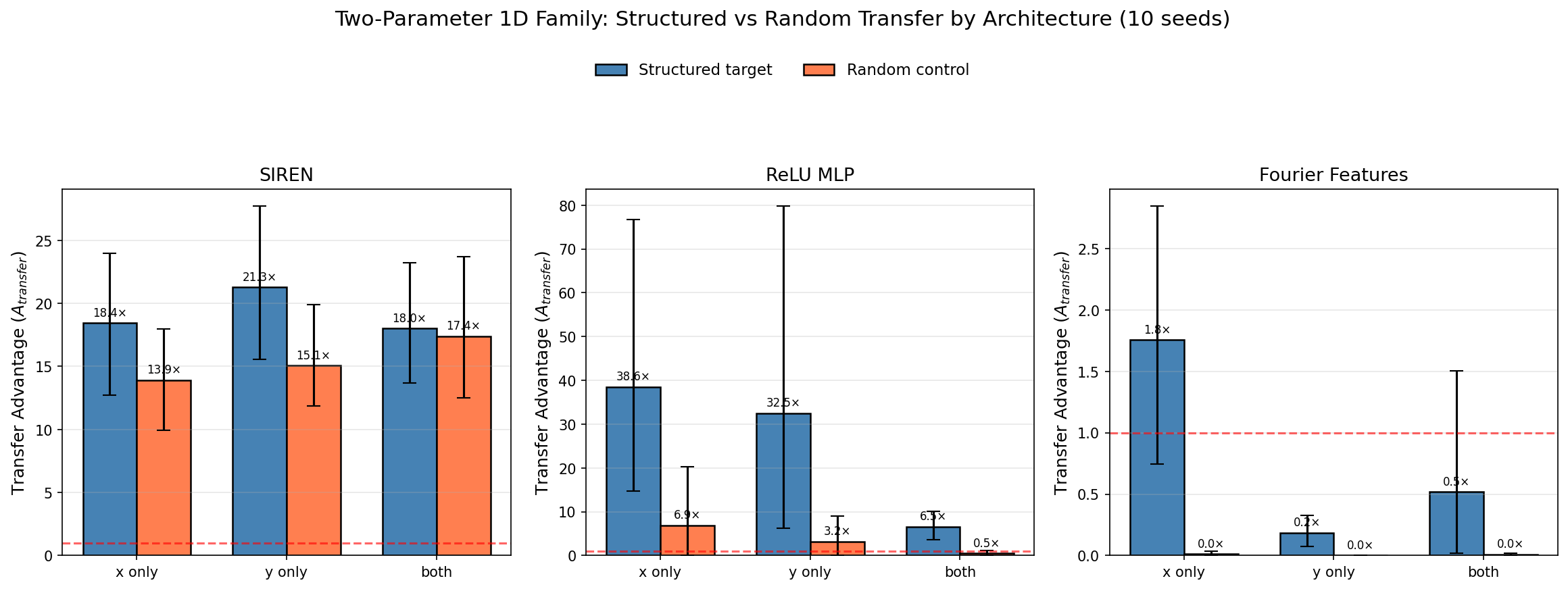}
\caption{Paired structured-versus-random transfer comparison on the controlled two-parameter 1D family across architectures over 10 seeds at the default architecture settings. Each panel shows mean transfer advantage, and error bars denote bootstrap 95\% confidence intervals for the mean. ReLU shows the clearest specificity, default SIREN remains weakly specific, and Fourier separates structured targets from random controls mainly because both transfers are small in absolute terms.}
\label{fig:2d_specificity}
\end{figure}

Figure~\ref{fig:2d_specificity} visualizes the same paired comparison on the controlled two-parameter 1D family. The clearest qualitative contrast is between ReLU, which separates structured targets from random controls in all three directions under the pre-specified one-sided tests, and SIREN, whose structured and random bars remain close.

\begin{figure}[t]
\centering
\includegraphics[width=0.5\textwidth]{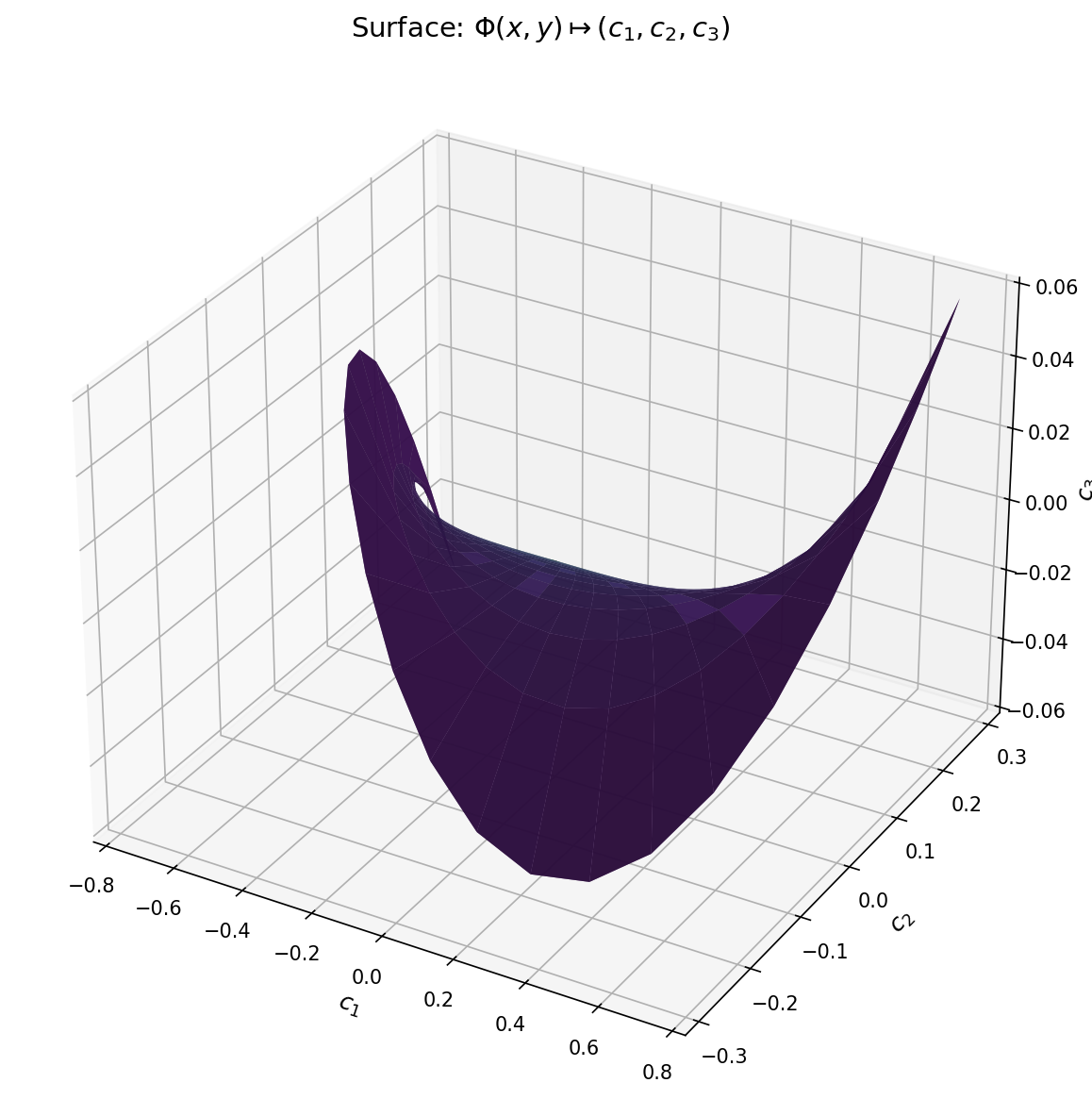}
\caption{Analytic coefficient surface $\Phi: (x,y) \mapsto (c_1, c_2, c_3)$ for the two-parameter 1D family. This surface is computed directly from the closed-form coefficients, not from neural network representations.}
\label{fig:2d_surface}
\end{figure}

The surface $\Phi(x,y) = (c_1, c_2, c_3)$ visualization (Figure~\ref{fig:2d_surface}) shows the \emph{analytic} coefficient map, not learned neural representations.

\begin{figure}[t]
\centering
\includegraphics[width=0.75\textwidth]{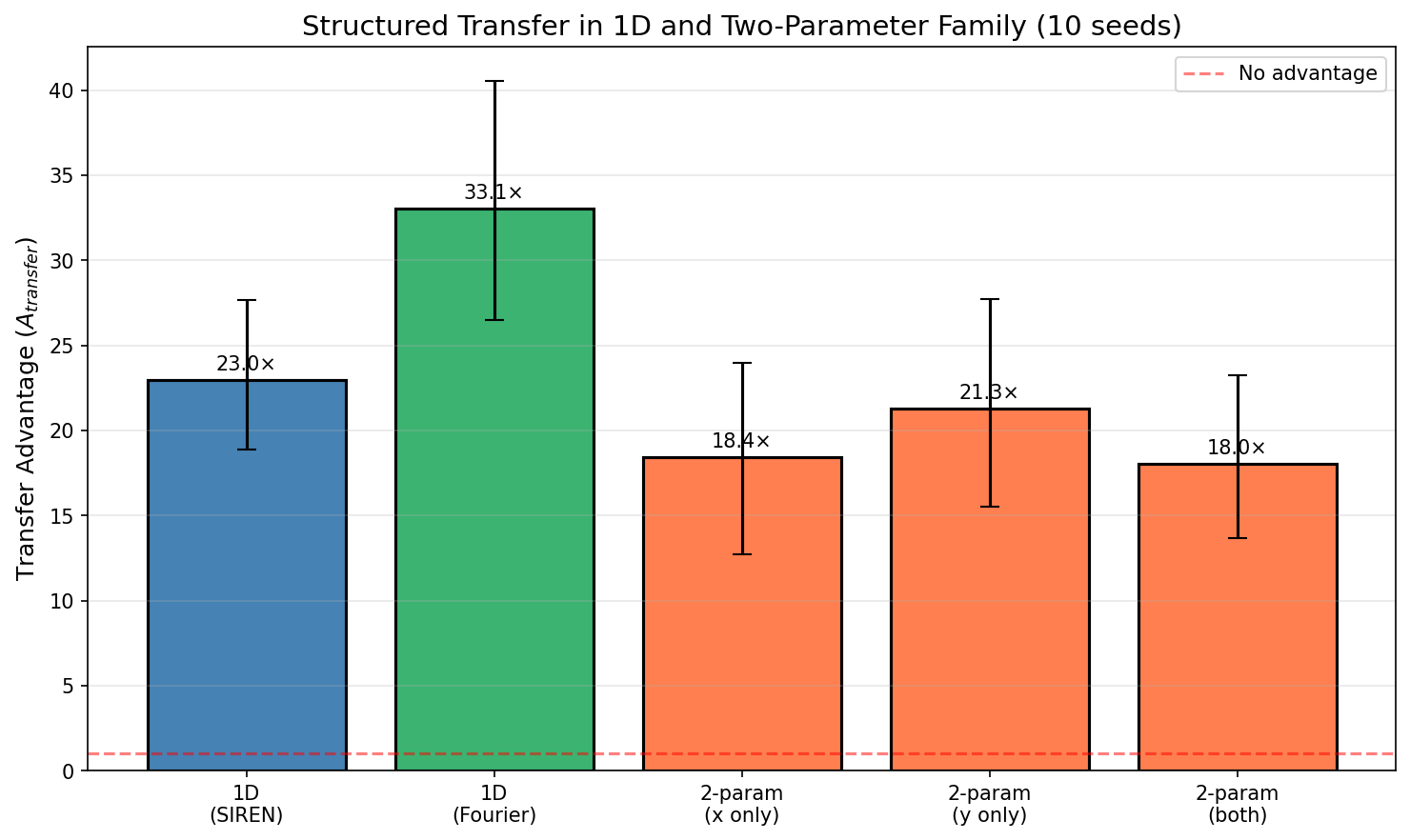}
\caption{Structured-target transfer comparison in the controlled 1D geometric test and the default SIREN sweep on the controlled two-parameter 1D family over 10 seeds. Bars show mean transfer advantage and error bars show bootstrap 95\% confidence intervals. Table~\ref{tab:2d_specificity} extends the comparison to ReLU and Fourier, revealing that the two-parameter architecture ranking differs qualitatively from the 1D ranking.}
\label{fig:2d_collapse}
\end{figure}

\subsection{Scaling Law: Negative Result Across the Implemented 1D Architectures}

We tested the heuristic prediction $A \propto 1/\Delta t^2$ in the implemented 1D setting for all three architectures, using $t_{\mathrm{source}}=0.5$, eight target offsets $\Delta t \in [0.01, 0.2]$, and three seeds averaged at each offset. The regression therefore remains a limited probe: it averages over only three seeds per architecture and uses one optimization protocol throughout. It nevertheless supports a clear qualitative conclusion. The fitted relations are:
\begin{itemize}
    \item \textbf{SIREN:} slope $0.026 \pm 0.010$, $R^2 = 0.53$, $H_0\!:\text{slope}=1$ rejected at $p = 8.5\times 10^{-11}$
    \item \textbf{ReLU MLP:} slope $0.308 \pm 0.032$, $R^2 = 0.94$, $H_0\!:\text{slope}=1$ rejected at $p = 6.1\times 10^{-7}$
    \item \textbf{Fourier Features:} slope $0.590 \pm 0.070$, $R^2 = 0.92$, $H_0\!:\text{slope}=1$ rejected at $p = 1.1\times 10^{-3}$
\end{itemize}

The usual \texttt{linregress} $p$-value tests the irrelevant null $H_0\!:\text{slope}=0$; the relevant comparison for this ansatz is $H_0\!:\text{slope}=1$. Under that test, the ansatz is rejected for all three architectures. The rejection does \emph{not} imply that transfer is independent of $\Delta t$: ReLU and Fourier both decay with $\Delta t$, but substantially more slowly than the predicted quadratic law. We therefore treat this as a negative empirical result against a simple heuristic model, not as a theorem about transfer scaling in INRs.

\begin{figure}[t]
\centering
\includegraphics[width=\textwidth]{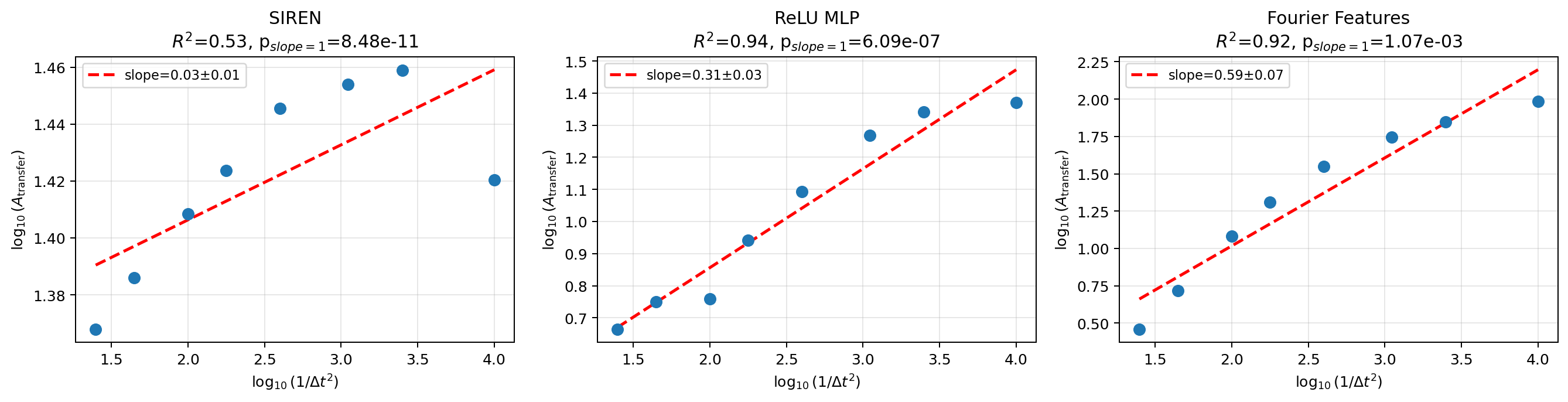}
\caption{Cross-architecture scaling-law audit of the prediction $A_{\text{transfer}} \propto 1/\Delta t^2$. Each panel plots $\log_{10} A$ versus $\log_{10}(1/\Delta t^2)$ for one architecture. The expected slope $1.0$ is rejected for SIREN, ReLU, and Fourier Features. ReLU and Fourier show clear decay with $\Delta t$, but much slower than the quadratic-law prediction.}
\label{fig:scaling}
\end{figure}

\section{PDE Benchmarks}

\subsection{Navier--Stokes Lid-Driven Cavity}

To assess whether the architecture-dependent transfer pattern established on controlled analytic families extends to a physically realistic setting, we apply the same INR transfer protocol to the steady lid-driven cavity problem. The target function family is parameterized by Reynolds number $Re \in \{100, 400, 1000\}$, which plays the role of the continuous parameter $t$ in the analytic tests. For each $Re$, we solve the steady incompressible Navier-Stokes equations
\[
  (\mathbf{u} \cdot \nabla)\mathbf{u} = -\nabla p + \tfrac{1}{Re}\nabla^2 \mathbf{u},
  \quad \nabla \cdot \mathbf{u} = 0,
\]
on $\Omega = [0,1]^2$ using a stream-function/vorticity finite-difference solver ($129 \times 129$ uniform grid), with the lid $u_x = 1$ at $y=1$ and no-slip on the remaining walls. The converged velocity field $\mathbf{u}(x,y;Re) = (u_x, u_y)$ serves as the INR regression target.
This is a steady, moderate-Reynolds-number laminar cavity benchmark, not a turbulent or time-dependent flow setting.

Each architecture maps $(x,y) \in [0,1]^2$ to $(u_x, u_y)$. All NS models use width 256 and depth 4; SIREN uses first-layer $\omega_0=30$, and the Fourier model uses 128 Fourier frequencies with $\sigma=10$. Training uses Adam at learning rate $10^{-4}$ throughout, with mini-batches of 2048 grid points sampled uniformly with replacement from the fixed $129 \times 129$ cavity grid. The transfer protocol matches the analytic tests only in broad structure: pretrain at source $Re_0$ for 50{,}000 gradient steps, then fine-tune at target $Re_1$ for 10{,}000 steps; scratch training also uses 10{,}000 steps. We test three directions $(Re_0, Re_1) \in \{(100,400),\,(400,1000),\,(100,1000)\}$; the remaining Reynolds number $Re_\perp$ serves as an alternate same-family source condition (pretrained at $Re_\perp$, fine-tuned at $Re_1$ with the same seed), not an independent random baseline. As a stronger null on the same target, we additionally copy the designated-source checkpoint, randomly shuffle all trainable weights while leaving fixed buffers such as Fourier $B$ unchanged, and then fine-tune this shuffled source to $Re_1$ with the same target budget. All experiments use $n=10$ seeds with bootstrap 95\% confidence intervals and paired \emph{two-sided} Wilcoxon signed-rank tests for designated-versus-alternate comparisons, plus one-sided paired Wilcoxon tests for designated-source transfer exceeding shuffled-source transfer. The NS experiment records per-seed transfer advantage in decibels,
$A_{\mathrm{dB}} = 10 \log_{10}(\mathrm{MSE}_{\mathrm{scratch}} / \mathrm{MSE}_{\mathrm{transfer}})$.
For comparability with the analytic tests, Table~\ref{tab:ns} reports the corresponding linear-ratio equivalents $10^{\bar A_{\mathrm{dB}}/10}$ obtained from the mean dB values, while the Wilcoxon tests are applied to the underlying per-seed dB values. Because the alternate-source condition is neither an independent random null nor a distance-matched control, the NS section should be interpreted as a source-choice comparison rather than as a structured-versus-random specificity test.

\subsection{Navier--Stokes Results}

\begin{figure}[t]
\centering
\includegraphics[width=0.9\textwidth]{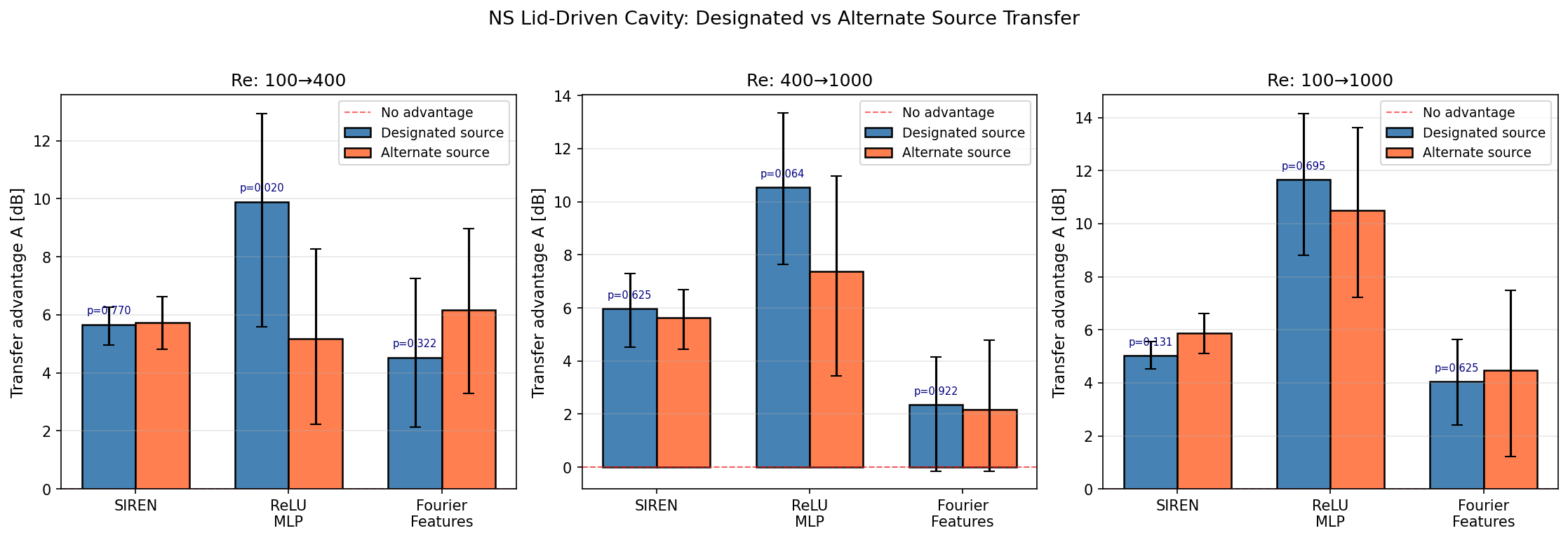}
\caption{NS benchmark: designated-source vs.\ alternate-source transfer advantage across architectures and Reynolds-number directions over 10 seeds, shown in the underlying dB metric used by the experiment. Error bars are bootstrap 95\% confidence intervals for the mean. ReLU shows the clearest separation between the two source conditions; SIREN transfers substantially but similarly from both sources; Fourier Features transfer weakly in both conditions.}
\label{fig:ns}
\end{figure}

Table~\ref{tab:ns} reports converted linear-ratio equivalents for the designated and alternate same-family sources, while Figure~\ref{fig:ns} shows the underlying dB values. Four patterns emerge.

\textbf{SIREN transfers substantially but shows little separation from the alternate-source condition.} Designated-source transfer ranges from $3.2\times$ to $4.0\times$ across all three directions, confirming that SIREN weights are reused across the Reynolds-number family. However, alternate-source transfer is similarly large ($3.7$--$3.9\times$), and none of the paired Wilcoxon tests reaches the $0.05$ threshold (all $p \geq 0.13$). This is consistent with the SIREN pattern from the analytic tests: weight reuse is real but not sharply selective.

\textbf{ReLU MLP is the most source-conditioned.} For $100\to400$, designated-source transfer ($9.8\times$, 95\% CI $[3.6,19.6]$) exceeds alternate-source transfer ($3.3\times$, CI $[1.7,6.7]$), with $p=0.020$. For $400\to1000$ the separation is similar ($11.4\times$ vs $5.5\times$) and borderline significant ($p=0.064$). For $100\to1000$, both designated-source ($14.7\times$) and alternate-source ($11.2\times$) transfer are large and comparable ($p=0.695$), suggesting that across the widest Reynolds gap the pretrained representation is useful regardless of which same-family source was used. Overall, ReLU provides the clearest separation between the designated and alternate source conditions on this benchmark.

\textbf{Fourier Features transfer weakly.} Designated-source transfer is modest ($1.7$--$2.8\times$) and never reliably exceeds alternate-source transfer (all $p \geq 0.32$). The CI for $400\to1000$ includes $1\times$, indicating that fine-tuning from a Fourier pretrain provides limited or unreliable benefit over training from scratch.

\textbf{The shuffled-weight source null confirms that the learned source weights usually matter.} Table~\ref{tab:ns-shuffled} compares designated-source transfer against a shuffled copy of the same source checkpoint. SIREN and ReLU designated-source transfer exceeds the shuffled source by $10.3$--$13.1$ dB across all Reynolds-number directions (all one-sided paired $p \leq 0.002$). Fourier is mixed: $100\to400$ still exceeds the shuffled source by $4.58$ dB ($p=0.0068$), but the $400\to1000$ and $100\to1000$ differences are smaller and not significant at $0.05$. Thus the alternate-source control understated one useful fact for SIREN: even when transfer is not selective among Reynolds-number sources, the learned source weights are still far better than a shuffled-weight null.

\begin{table}[t]
\centering
\small
\begin{tabular}{llccl}
\hline
Architecture & Direction & Designated-source transfer & Alternate-source transfer & $p$ \\
\hline
\multirow{3}{*}{SIREN}
  & $100\to400$  & $3.67\times$ $[3.13,\,4.23]$ & $3.74\times$ $[3.02,\,4.59]$ & $0.770$ \\
  & $400\to1000$ & $3.97\times$ $[2.83,\,5.37]$ & $3.66\times$ $[2.78,\,4.66]$ & $0.625$ \\
  & $100\to1000$ & $3.19\times$ $[2.84,\,3.60]$ & $3.87\times$ $[3.24,\,4.58]$ & $0.131$ \\
\hline
\multirow{3}{*}{ReLU MLP}
  & $100\to400$  & $9.76\times$  $[3.61,\,19.65]$ & $3.30\times$ $[1.67,\,6.70]$  & $0.020^*$ \\
  & $400\to1000$ & $11.36\times$ $[5.79,\,21.67]$ & $5.46\times$ $[2.20,\,12.51]$ & $0.064$ \\
  & $100\to1000$ & $14.66\times$ $[7.59,\,26.02]$ & $11.22\times$ $[5.27,\,23.04]$ & $0.695$ \\
\hline
\multirow{3}{*}{Fourier Features}
  & $100\to400$  & $2.83\times$ $[1.63,\,5.32]$ & $4.13\times$ $[2.13,\,7.87]$ & $0.322$ \\
  & $400\to1000$ & $1.73\times$ $[0.97,\,2.60]$ & $1.64\times$ $[0.97,\,3.02]$ & $0.922$ \\
  & $100\to1000$ & $2.54\times$ $[1.74,\,3.67]$ & $2.80\times$ $[1.33,\,5.62]$ & $0.625$ \\
\hline
\end{tabular}
\caption{10-seed Navier-Stokes benchmark (lid-driven cavity, $Re \in \{100,400,1000\}$). The underlying experiment records per-seed transfer advantage in dB, $A_{\mathrm{dB}} = 10 \log_{10}(\mathrm{MSE}_{\mathrm{scratch}}/\mathrm{MSE}_{\mathrm{transfer}})$. For comparability with the analytic tests, each table entry reports the linear-ratio equivalent of the mean dB value, $10^{\bar A_{\mathrm{dB}}/10}$, rather than the mean of per-seed linear ratios. The alternate-source condition pretrains on $Re_\perp$, the third Reynolds number in $\{100,400,1000\}$; it is not an independent random baseline. The Wilcoxon $p$-value is the paired two-sided test on the per-seed dB values comparing designated-source and alternate-source transfer. $^*p < 0.05$.}
\label{tab:ns}
\end{table}

\begin{table}[t]
\centering
\small
\resizebox{\textwidth}{!}{%
\begin{tabular}{llcccc}
\hline
Architecture & Direction & Designated-source $A_{\mathrm{dB}}$ & Shuffled-source $A_{\mathrm{dB}}$ & Difference & $p$ \\
\hline
\multirow{3}{*}{SIREN}
  & $100\to400$  & $5.65$ $[4.96,\,6.26]$ & $-4.86$ $[-5.56,\,-4.19]$ & $10.51$ $[9.60,\,11.35]$ & $0.0010$ \\
  & $400\to1000$ & $5.99$ $[4.52,\,7.30]$ & $-5.64$ $[-6.26,\,-5.02]$ & $11.63$ $[10.34,\,12.89]$ & $0.0010$ \\
  & $100\to1000$ & $5.03$ $[4.54,\,5.57]$ & $-5.28$ $[-5.73,\,-4.92]$ & $10.32$ $[9.65,\,11.06]$ & $0.0010$ \\
\hline
\multirow{3}{*}{ReLU MLP}
  & $100\to400$  & $9.90$ $[5.58,\,12.93]$ & $-3.22$ $[-4.46,\,-1.76]$ & $13.12$ $[9.26,\,15.45]$ & $0.0020$ \\
  & $400\to1000$ & $10.55$ $[7.63,\,13.36]$ & $-2.09$ $[-3.12,\,-0.89]$ & $12.64$ $[9.66,\,15.11]$ & $0.0010$ \\
  & $100\to1000$ & $11.66$ $[8.80,\,14.15]$ & $0.12$ $[-1.32,\,1.72]$ & $11.54$ $[9.48,\,13.18]$ & $0.0010$ \\
\hline
\multirow{3}{*}{Fourier Features}
  & $100\to400$  & $4.52$ $[2.12,\,7.26]$ & $-0.05$ $[-3.09,\,2.96]$ & $4.58$ $[2.01,\,7.31]$ & $0.0068$ \\
  & $400\to1000$ & $2.37$ $[-0.14,\,4.15]$ & $1.48$ $[-2.08,\,5.11]$ & $0.89$ $[-1.87,\,3.41]$ & $0.2783$ \\
  & $100\to1000$ & $4.05$ $[2.41,\,5.65]$ & $2.21$ $[0.80,\,3.62]$ & $1.84$ $[-0.11,\,3.86]$ & $0.0801$ \\
\hline
\end{tabular}%
}
\caption{NS shuffled-weight source control in the underlying dB metric. The shuffled source is made by copying the designated-source checkpoint, randomly permuting trainable weights, leaving fixed buffers such as Fourier $B$ unchanged, and fine-tuning to the same target with the same budget. The $p$-value is the one-sided paired Wilcoxon test for designated-source transfer exceeding shuffled-source transfer.}
\label{tab:ns-shuffled}
\end{table}

\subsection{Relation to the Analytic Tests}

The NS benchmark is broadly consistent with the two-parameter 1D analytic family rather than a literal replication of it, but its control is different. It compares transfer from a designated source Reynolds number against transfer from the third same-family source rather than against an independent random null. Interpreted that way, ReLU again shows the clearest separation between source conditions, SIREN remains broadly reusable, and Fourier remains weak in absolute transfer. The shuffled-weight null adds a second interpretation: SIREN is broad across Reynolds-number sources, but not merely generic initialization reuse, because its designated-source transfer exceeds shuffled-source transfer by more than $10$ dB in all three directions. ReLU is both source-conditioned and strongly above the shuffled null. Fourier remains the weakest and most variable case. The weak NS Fourier result at $\sigma=10$ is at least qualitatively compatible with the analytic-test sigma sweep, which shows that the default analytic setting $\sigma=10$ is poor while smaller values can perform much better. That said, the sigma ablation was run only on the analytic family, so it does not by itself determine the optimal NS bandwidth.

\subsection{1D PDE Reference-Solution Suite}

Table~\ref{tab:pde-suite} summarizes the full-budget heat, Burgers, and NLS experiments. Each entry aggregates over two target directions and ten seeds ($n=20$ terminal transfer ratios). The control condition is the alternate same-family source, not an independent random target, so these values should be read as source-conditioned transfer comparisons rather than structured-versus-random specificity tests.

\begin{table}[t]
\centering
\small
\resizebox{\textwidth}{!}{%
\begin{tabular}{llcccc}
\hline
PDE & Architecture & Designated-source transfer & Alternate-source transfer & Specificity ratio & $p$ \\
\hline
Heat & SIREN & $10.83\times$ $[10.16,\,11.50]$ & $3.66\times$ $[3.32,\,4.01]$ & $3.15\times$ $[2.70,\,3.63]$ & $9.5{\times}10^{-7}$ \\
Heat & ReLU MLP & $2.14\times$ $[1.64,\,2.81]$ & $0.32\times$ $[0.25,\,0.39]$ & $7.15\times$ $[5.87,\,8.46]$ & $9.5{\times}10^{-7}$ \\
Heat & Fourier Features & $1.38\times$ $[1.03,\,1.78]$ & $0.18\times$ $[0.14,\,0.22]$ & $8.20\times$ $[7.05,\,9.28]$ & $9.5{\times}10^{-7}$ \\
\hline
Burgers & SIREN & $8.48\times$ $[6.23,\,10.79]$ & $3.14\times$ $[2.58,\,3.70]$ & $2.51\times$ $[2.19,\,2.83]$ & $9.5{\times}10^{-7}$ \\
Burgers & ReLU MLP & $4.76\times$ $[3.41,\,6.35]$ & $1.49\times$ $[0.87,\,2.20]$ & $5.08\times$ $[4.00,\,6.18]$ & $9.5{\times}10^{-7}$ \\
Burgers & Fourier Features & $19.09\times$ $[15.15,\,23.69]$ & $3.25\times$ $[2.70,\,3.92]$ & $6.29\times$ $[5.05,\,7.63]$ & $9.5{\times}10^{-7}$ \\
\hline
NLS & SIREN & $20.05\times$ $[18.51,\,21.60]$ & $17.79\times$ $[16.53,\,19.27]$ & $1.13\times$ $[1.08,\,1.19]$ & $5.1{\times}10^{-4}$ \\
NLS & ReLU MLP & $4.73\times$ $[3.39,\,6.38]$ & $1.64\times$ $[1.08,\,2.42]$ & $3.51\times$ $[2.93,\,4.15]$ & $9.5{\times}10^{-7}$ \\
NLS & Fourier Features & $19.50\times$ $[15.14,\,24.06]$ & $6.94\times$ $[5.17,\,9.16]$ & $3.35\times$ $[2.59,\,4.28]$ & $1.9{\times}10^{-6}$ \\
\hline
\end{tabular}%
}
\caption{Full-budget 1D PDE reference-solution benchmarks over heat, viscous Burgers, and focusing cubic NLS. Values are macro means over two target directions and ten seeds, with bootstrap 95\% confidence intervals. Transfer advantage is the full-grid terminal ratio $\mathrm{MSE}_{\mathrm{scratch}}/\mathrm{MSE}_{\mathrm{transfer}}$. The alternate-source condition pretrains on the opposite target parameter within the same PDE family; it is not an independent random null. The $p$-value is the one-sided paired Wilcoxon signed-rank test comparing designated-source transfer to alternate-source transfer over the 20 paired direction-seed values.}
\label{tab:pde-suite}
\end{table}

The PDE benchmarks add two useful checks to the analytic-test and NS story. First, the architecture ranking is not universal across equations: SIREN has the largest heat transfer magnitude, Fourier Features dominate Burgers, and SIREN and Fourier are comparable in NLS magnitude. Second, magnitude and specificity remain separable. In NLS, SIREN reaches $20.05\times$ designated-source transfer, but alternate-source transfer is also $17.79\times$, giving only a $1.13\times$ specificity ratio. ReLU has lower NLS magnitude ($4.73\times$) but a cleaner $3.51\times$ ratio, while Fourier combines high NLS magnitude ($19.50\times$) with stronger separation from the alternate source ($3.35\times$). Thus the complex-valued dispersive PDE does not collapse the architecture dependence; it makes the distinction between weight reuse and source specificity more visible.

The auxiliary controls clarify what the terminal ratios do and do not mean. When scratch training uses the same low learning rate as fine-tuning, absolute transfer ratios become much larger---for example, matched-optimizer designated-source ratios reach $250.9\times$ for heat/SIREN, $353.3\times$ for Burgers/ReLU, and $1974.0\times$ for NLS/Fourier. This confirms that terminal transfer magnitude is strongly schedule-sensitive. However, the designated-versus-alternate specificity ratios are unchanged by this common scratch numerator, so the source-specific conclusions in Table~\ref{tab:pde-suite} do not rely on the faster scratch learning rate. We therefore keep the fixed-budget terminal metric in the main tables because it answers the operational question ``which initialization gives lower loss after the same target budget?'', while the matched-optimizer results are interpreted as causal sensitivity checks rather than as replacements for the headline warm-start protocol. The Fourier feature-map ablation is more diagnostic: copying the pretrained MLP weights but resampling $B$ gives only $0.010$--$0.011\times$ transfer under the default scratch baseline, and reusing only $B$ with random MLP weights gives only $0.011$--$0.023\times$. Under the matched-optimizer scratch baseline these variants rise to roughly scratch-level performance ($0.62$--$1.36\times$), but remain far below the full Fourier transfer. Thus Fourier transfer in these PDE runs is not explained by the fixed random projection alone; it depends on the coupled reuse of the projection and the trained weights.

\section{Discussion}

\subsection{Architecture Controls Magnitude and Specificity}

The controlled 1D geometric sweep shows that two different quantities must be distinguished: \emph{how much} transfer occurs, and \emph{what that transfer means}. Fourier Features have the largest mean structured transfer in the current 10-seed 1D sweep, which is consistent with the hypothesis that explicit frequency encoding is well matched to the spectral structure of $g_t(x) = \sqrt{x^2 + t^2}$. But specificity is a separate issue. SIREN still shows substantial transfer on independent-seed random targets, ReLU remains close to zero on the random control, and Fourier retains nonzero random transfer that is nonetheless much smaller than its structured transfer.

The two-parameter sweep sharpens this point by showing that there is no single universal architecture ranking. At the default settings, ReLU, not Fourier, shows the clearest structured-vs-random separation. Fourier still separates structured targets from random controls, but mainly because both transfers are small in absolute terms. Default SIREN remains reusable, but that reuse is only weakly specific. The fairness sweeps show that this is not a rigid family-level property: smaller SIREN initialization scale and, more moderately, larger $\omega_0$ can restore clear specificity, whereas ReLU remains the most robust family across the tested width, depth, and fine-tuning-learning-rate sweeps. Architecture therefore affects both weight reuse and the degree to which that reuse reflects genuine target-family continuity, and these effects depend on both the dimensionality of the parameter family and the operating point inside each architecture family.

The Navier-Stokes benchmark shows that this two-parameter architecture picture is not obviously an artifact of the controlled cosine family. On a physically realistic steady-flow system with a one-dimensional parameter ($Re$), the qualitative pattern is similar under the alternate-source control: ReLU is the clearest discriminator between source conditions, SIREN is broadly reusable from either source, and Fourier Features provide weak transfer in both conditions. The shuffled-weight null strengthens this interpretation by showing that SIREN and ReLU transfer are not reproduced by destroying the learned source weights; the Fourier result remains more ambiguous in two of the three Reynolds-number directions. The NS evidence is still limited to one steady laminar geometry, but it is no longer based only on the alternate-source comparison.

\subsection{Proper Null Models Matter}

Independent-seed random controls are essential for assessing transfer specificity. In the 10-seed controlled 1D geometric test, the effect of this null model is not universal across architectures. For ReLU, independent-seed random controls drive transfer close to zero. For Fourier, random-control transfer remains far below structured-target transfer but is no longer negligible in mean. For SIREN, substantial random transfer survives. The correct conclusion is therefore not that ``random transfer vanishes,'' but that null-model separation itself is architecture-dependent.

\subsection{What the Two-Parameter Result Actually Says}

The two-parameter experiment supports a more structured statement than either a universal success or a universal failure story. At the default settings, ReLU shows clear specificity under the independent-seed null model, default SIREN shows substantial but weakly specific transfer, and Fourier shows poor absolute transfer despite statistically significant structured-vs-random separation. The fairness sweeps narrow the claim further: the main table should be interpreted as a default-configuration comparison, not as a statement that every SIREN setting is weakly specific or that every large terminal ratio inside the ReLU family represents a robust improvement. Smaller SIREN initialization scale can restore strong specificity, while some apparently extreme ReLU wins are largely terminal-ratio outliers and look much more modest under the curve-aware summary. The two-parameter result is therefore not ``transfer collapses in higher-dimensional parameter space,'' nor ``all architectures exploit parameter continuity equally well.'' It is that the nature of transfer becomes strongly architecture-dependent and qualitatively different from the 1D case.

A coarse 10-seed Fourier-feature-scale ablation with independent random controls shows that the weak two-parameter Fourier result is strongly hyperparameter-dependent. With the default scale $\sigma = 10$, the macro-mean structured transfer is only $0.59\times$ and the macro-mean random transfer is $0.08\times$. Lowering the scale to $\sigma = 1$ raises the macro-mean structured transfer to $50.0\times$ while keeping macro-mean random transfer at $6.3\times$; $\sigma = 2$ gives $39.2\times$ and $3.1\times$. By contrast, $\sigma = 0.5$ produces even larger structured transfer ($63.0\times$) but also very large random-control transfer ($49.0\times$), so specificity deteriorates sharply. The paired macro structured-target comparisons versus $\sigma = 10$ are significant for $\sigma \in \{0.5,1,2\}$ (all $p = 0.00195$). The revised ablation therefore suggests a genuine magnitude-specificity tradeoff inside the analytic Fourier family: the default $\sigma = 10$ is a poor operating point for this controlled analytic problem, but smaller $\sigma$ is not uniformly better. This analytic-only ablation does not by itself determine the optimal NS bandwidth, because the NS benchmark differs in input dimensionality, feature count, width, and control condition.

\subsection{Efficacy, Cost, and Practical Guidance}

The transfer ratios reported here measure fixed-budget fine-tuning benefit, not total wall-clock superiority over training from scratch. For the analytic and 1D PDE studies, a source model is pretrained for 1500 steps and then fine-tuned for 300 steps, while scratch training for a target also uses 300 steps. Thus, if a source model is used for only one target, the pretraining cost is not amortized. The relevant use case is a parametric sweep in which a trained source representation can seed multiple nearby targets, or a production workflow in which pretrained INR surrogates are updated repeatedly as parameters change. In the completed heat/Burgers/NLS run, the full 10-seed, three-architecture PDE benchmark suite took about $5.0\times10^3$ seconds end-to-end, with cached reference solutions reused across reruns; the additional matched-optimizer, Fourier-ablation, and saved-curve PDE controls increase output size and target-phase training work but reuse the same cached reference solutions. The NS benchmark is more expensive by design, using 50{,}000 pretraining steps and 10{,}000 fine-tuning or scratch steps per Reynolds-number direction.

The practical architecture guidance is correspondingly conditional. If the workflow values \emph{selective} reuse across related parameters, ReLU MLPs are the most robust default in our experiments: they give the clearest random-control separation in the controlled families and the clearest designated-source advantage in NS. If the workflow values broad warm-start reuse and can tolerate weaker source specificity, SIREN can be useful, but its strong transfer on random or alternate sources means that transfer magnitude alone should not be interpreted as evidence of physical parameter continuity. Fourier features require explicit bandwidth tuning: they are strong on the 1D geometric test and Burgers, weak at the default scale on the two-parameter analytic family and NS, and substantially improved by smaller frequency scales in the analytic sigma sweep. For computational-physics use, the main recommendation is therefore to report source-specific controls, seed variability, and architecture hyperparameters together with any INR transfer gains.

\subsection{Revised Theoretical Picture}

Our results suggest the following revised picture:
\begin{enumerate}
    \item Transfer performance depends strongly on architecture, and transfer specificity depends on architecture even more strongly.
    \item The two-parameter architecture ranking differs qualitatively from the 1D ranking at the default settings: ReLU is the clearest discriminator, default SIREN is weakly specific, and Fourier has poor absolute transfer. Fairness sweeps weaken the strongest family-level version of this claim by showing that tuned SIREN can also become clearly specific.
    \item A broadly similar architecture pattern appears on a physically realistic PDE benchmark. On the Navier-Stokes lid-driven cavity family ($Re \in \{100,400,1000\}$) under an alternate same-family source control, ReLU again shows the clearest source-conditioned advantage, SIREN again transfers broadly, and Fourier again transfers weakly. A shuffled-weight null shows that SIREN/ReLU transfer is nevertheless strongly above destroyed-weight reuse.
    \item The heat, Burgers, and NLS reference-solution benchmarks broaden the evidence but do not produce a single architecture winner: transfer magnitude and alternate-source specificity vary by equation, with the complex-valued NLS case showing especially clear separation between broad SIREN reuse and more selective ReLU/Fourier transfer.
    \item Matched-optimizer and Fourier feature-map controls show that terminal transfer magnitudes are optimizer-schedule sensitive, while Fourier transfer depends on coupled reuse of the fixed projection matrix and trained MLP weights rather than on either component alone.
    \item Static diagnostics remain weakly informative: PR and Hessian sharpness do not separate the targets, and independent-seed CKA is not specificity-selective.
    \item Simple theoretical models do not capture the observed transfer dynamics: the slope-$1$ scaling ansatz is rejected for all three implemented 1D architectures.
\end{enumerate}

\section{Conclusion}

This paper studies transfer in implicit neural representations as a property of the representation itself, not only as a convenience for faster fine-tuning. Across controlled analytic families, Navier--Stokes, and 1D PDE reference-solution benchmarks, the main result is consistent: transfer magnitude and transfer specificity are separate quantities, and both depend strongly on architecture. Fourier Features are strongest on the controlled 1D geometric test, ReLU is the clearest discriminator on the controlled two-parameter family and Navier--Stokes benchmark, and the heat/Burgers/NLS suite shows that no single architecture dominates every equation in absolute transfer.

The broader lesson is methodological. Transfer gains should be interpreted together with explicit control conditions, because large warm-start improvements can coexist with weak source specificity. In our experiments, SIREN often reuses weights broadly, ReLU is usually more selective, and Fourier Features can swing from strong to weak depending on bandwidth. Static diagnostics such as participation ratio, Hessian sharpness, and independent-seed CKA do not provide a reliable shortcut, and the simple scaling law $A_{\text{transfer}} \propto 1/\Delta t^2$ does not describe the observed dynamics. For readers interested in coordinate networks and scientific machine learning, the practical message is that architecture choice should be evaluated as a transfer-learning design decision, not only as a single-task approximation choice.

\paragraph{Limitations.} The study is still bounded by a fixed optimization protocol, 10-seed experiments in the main suites, and relatively small hyperparameter sweeps. The analytic tests use independent random controls, but the PDE benchmarks rely on alternate-source or shuffled-weight controls rather than fully off-family nulls. We also do not yet mirror the Fourier feature-map reuse ablation from the PDE suite in the analytic headline experiments, so the analytic Fourier results still combine MLP-weight reuse with reuse of the fixed random feature map. The Navier--Stokes study covers one steady laminar geometry and also uses larger models and a different schedule than the analytic suites, so it should be read as a qualitative replication rather than a fully harmonized cross-setting comparison. The 1D PDE suite uses supervised regression on fixed periodic grids with only two target parameters per equation. Absolute transfer magnitudes are also sensitive to learning-rate choices and, for Fourier Features, to bandwidth and feature-map reuse. These limitations do not overturn the central architecture-dependent patterns, but they do bound the claims to the tested architectures, controls, and transfer regimes.

\section*{Declaration of competing interest}
The author declares no competing financial or non-financial interests in relation to the work described.

\section*{Declaration of generative AI and AI-assisted technologies in the manuscript preparation process}
During the preparation of this work, the author used ChatGPT and Claude in order to assist with code drafting and debugging for experimental workflows, and with language polishing during manuscript preparation. After using these services, the author reviewed and edited the content as needed and takes full responsibility for the content of the published article.

\bibliographystyle{plainnat}
\bibliography{main}

\end{document}